\documentclass[journal,twoside,web]{ieeecolor}
\usepackage{generic}
\usepackage{cite}
\usepackage{amsmath,amssymb,amsfonts}
\usepackage{algorithmic}
\usepackage{graphicx}
\usepackage{textcomp}
\usepackage{subcaption}
\usepackage{caption}
\usepackage{multirow}
\usepackage{booktabs}
\usepackage{makecell}
\usepackage{mathrsfs}
\usepackage{bbm}

\def\BibTeX{{\rm B\kern-.05em{\sc i\kern-.025em b}\kern-.08em
    T\kern-.1667em\lower.7ex\hbox{E}\kern-.125emX}}

\begin{document}
\title{A Chebyshev Confidence Guided Source-Free Domain Adaptation Framework for Medical Image Segmentation}

\author{Jiesi Hu, Yanwu Yang, Xutao Guo, Jinghua Wang\textsuperscript{*}, Ting Ma\textsuperscript{*}
\thanks{This work was supported in part by grants from the National Natural Science Foundation of P.R. China (62276081, 62106113), Innovation Team and Talents Cultivation Program of National Administration of Traditional Chinese Medicine (NO:ZYYCXTD-C-202004), Basic Research Foundation of Shenzhen Science and Technology Stable Support Program (GXWD20201230155427003-20200822115709001) and The Major Key Project of PCL (PCL2021A06). (Corresponding author: Jinghua Wang, Ting Ma.)}
\thanks{Jiesi Hu , Yanwu Yang, and Xutao Guo are with School of Electronics and Information Engineering,
Harbin Institute of Technology at Shenzhen, and The Peng Cheng Laboratory. 
(e-mail: 405323011@qq.com, yangyanwu1111@gmail.com, 18B952052@stu.hit.edu.cn)}
\thanks{Jinghua Wang is with School of Computer Science and Technology, Harbin Institute of Technology at Shenzhen. (e-mail: wangjinghua@hit.edu.cn)}
\thanks{Ting Ma is with School of Electronics and Information Engineering,
Harbin Institute of Technology at Shenzhen, The Peng Cheng Laboratory, Guangdong Provincial Key Laboratory of Aerospace Communication and Networking Technology, Harbin Institute of Technology, Shenzhen, and
International Research Institute for Artifcial Intelligence, Harbin Institute of Technology, Shenzhen.
(e-mail: tma@hit.edu.cn)}
}

\maketitle

\begin{abstract}
Source-free domain adaptation (SFDA) aims to adapt models trained on a labeled source domain to an unlabeled target domain without the access to source data. In medical imaging scenarios, the practical significance of SFDA methods has been emphasized due to privacy concerns. Recent State-of-the-art SFDA methods primarily rely on self-training based on pseudo-labels (PLs). Unfortunately, PLs suffer from accuracy deterioration caused by domain shift, and thus limit the effectiveness of the adaptation process. To address this issue, we propose a Chebyshev confidence guided SFDA framework to accurately assess the reliability of PLs and generate self-improving PLs for self-training. The Chebyshev confidence is estimated by calculating probability lower bound of the PL confidence, given the prediction and the corresponding uncertainty. Leveraging the Chebyshev confidence, we introduce two confidence-guided denoising methods: direct denoising and prototypical denoising. Additionally, we propose a novel teacher-student joint training scheme (TJTS) that incorporates a confidence weighting module to improve PLs iteratively. The TJTS, in collaboration with the denoising methods, effectively prevents the propagation of noise and enhances the accuracy of PLs. Extensive experiments in diverse domain scenarios validate the effectiveness of our proposed framework and establish its superiority over state-of-the-art SFDA methods. Our paper contributes to the field of SFDA by providing a novel approach for precisely estimating the reliability of pseudo-labels and a framework for obtaining high-quality PLs, resulting in improved adaptation performance.
\end{abstract}


\section{Introduction}
\label{sec:introduction}
Deep neural network (DNN) models have achieved remarkable success in a wide range of visual recognition tasks \cite{Dosovitskiy2020an,he2016deep,ronneberger2015u,chen2017deeplab}. However, these models often suffer significant performance degradation when confronted with distribution or domain shifts which often exist between the training (source) and test (target) domains \cite{liu2021cycle,liu2023memory}. This issue is particularly prevalent in medical imaging, where a model trained on data from one clinical centre may exhibit poor performance when applied to data from another clinical centre.

To overcome the challenges induced by domain shifts, numerous algorithms have been developed in the field of Unsupervised Domain Adaptation (UDA) \cite{gu2020spherical,hoffman2018cycada,wang2019boundary}. 
However, most UDA techniques assume the availability of labeled source domain data for adaptation, which is often impractical and restricted in clinical applications due to privacy and security concerns \cite{yang2021federated}. Consequently, source-free domain adaptation (SFDA) has gained significant interest, particularly in medical image analysis \cite{kundu2020universal, liang2020we,bateson2020source,chen2021source, yang2022source,liu2023memory}. SFDA focuses on adapting a well-trained model to unlabeled data in the target domain, solely relying on the availability of a pre-trained model in the source domain. This makes SFDA a practical and efficient approach, as it allows clinical centers to adapt models to their own data without exchanging sensitive health information.

\begin{figure}
  \centering
  \includegraphics[width=\linewidth]{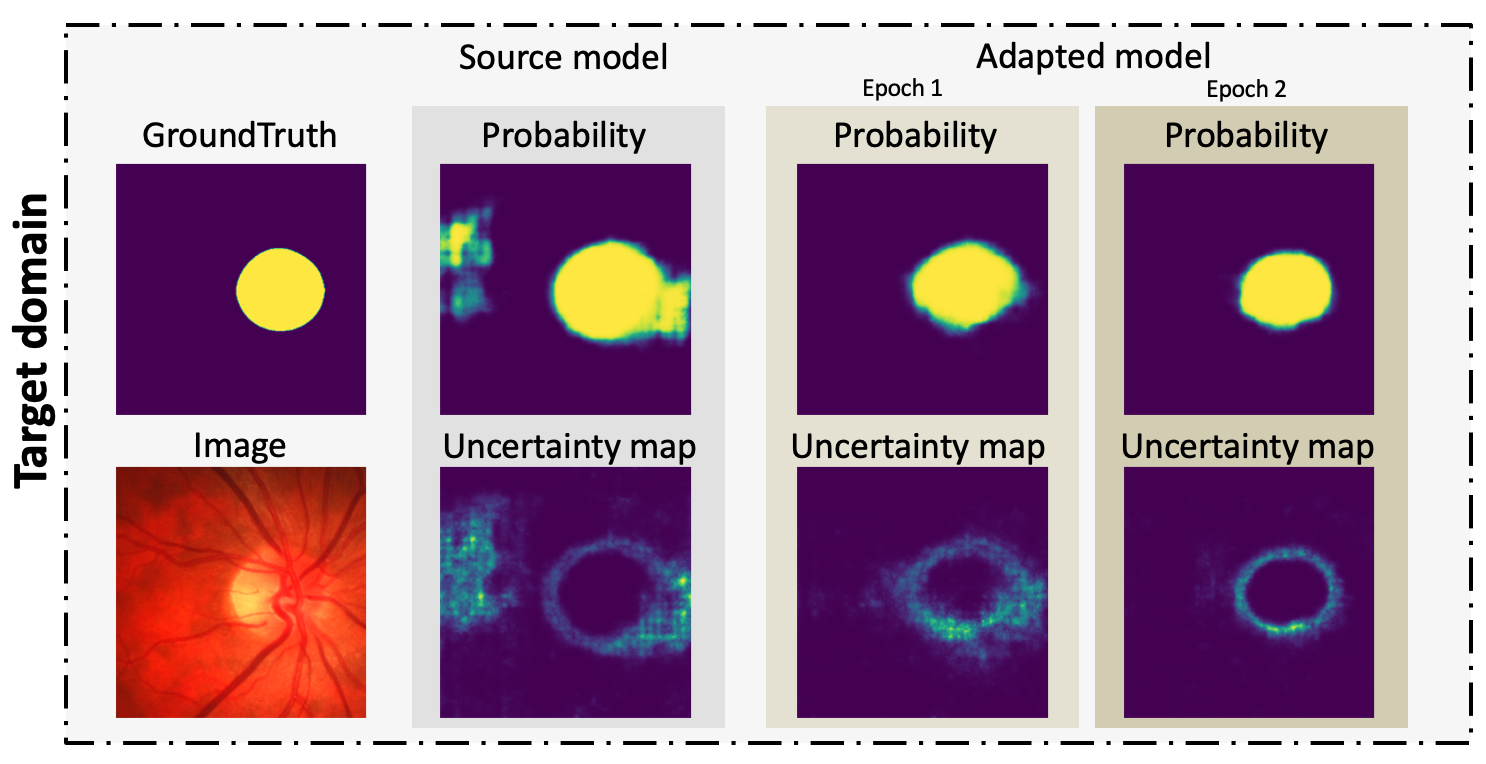}
  \captionsetup{font=scriptsize}
  \caption{
    The figure illustrates the evolution of segmentor outputs during the adaptation process. The outputs from the source model exhibit noticeable noise, which should be minimized in self-training. In contrast, the adapted model demonstrates improvement in predictions and reduced uncertainty. Leveraging the predictions from the adapted model enables the generation of more accurate pseudo-labels. Moreover, the probabilities and uncertainties offer insights into the model's reliability in various aspects of prediction. It is crucial to effectively combine these measures to determine the model's confidence in the pseudo-labels, which is a primary objective of this study.
    }
  \label{fig:01}
\end{figure}

Current SFDA methods employ various techniques, such as leveraging batch normalization parameters, contrastive learning, and target-to-source data transformation \cite{liang2020we,chen2021source,yang2021generalized,chen2022contrastive,liu2023memory}. Among these methods, the widely utilized approach is self-training guided by pseudo-labels (PLs) which are obtained by feeding the unlabeled target data to the model trained on labeled source data \cite{rizve2021defense}. However, during the early stages of adaptation, the PLs can be highly misleading or noisy, as shown in Fig. \ref{fig:01}. Using such PLs can propagate erroneous knowledge and hinder the effectiveness of domain adaptation. Therefore, obtaining accurate and noise-free PLs for self-training is crucial and requires proper attention \cite{chen2021source,karim2023c}.

The PLs-based methods have two major challenges. 
Firstly, the PLs may deviate significantly from the ground truth in the target domain due to domain shift. The inaccurate PLs hamper the model to achieve precise segmentation in the target domain. Consequently, there is an urgent need to generate more accurate PLs. Secondly, the PLs are normally noisy and an effective denoising method is crucial in the self-training process. Typically, researchers employ probability\cite{liu2023memory}, entropy \cite{liu2021cycle} or uncertainty \cite{chen2021source,karim2023c} to estimate the reliability of the PLs and eliminate unreliable ones. However, these methods fail to simultaneously consider the predicted probability and uncertainty, which are both crucial statistics for assessing the reliability of pseudo-labels. A more comprehensive and accurate method is required to assess the reliability of PLs. These two challenges are particularly critical in the field of medical imaging where the model is required to be highly stable and accurate. Designing a SFDA framework that generates high-quality PLs is a challenging task and the primary focus of this study. 

Our research focuses on improving the denoising and accuracy of PLs. To achieve this objective, we introduce a Chebyshev confidence guided SFDA framework that incorporates our proposed Chebyshev confidence estimation as a fundamental component. This estimation method calculates the probability lower bound of the PL confidence, taking into account the impact of uncertainty. For PL denoising, we employ two methods: direct denoising and prototypical denoising, both incorporating the proposed Chebyshev confidence. In direct denoising, we  identify and remove pixels with low confidence, by using Chebyshev confidence as an indicator. On the other hand, prototypical denoising accurately estimates prototypes and eliminates PLs with poor consistency based on Chebyshev confidence. By combining these complementary denoising methods, we effectively obtain clean PLs. We also propose a teacher-student joint training scheme to facilitate the knowledge of both the student and teacher models and ensure iterative improving of the PLs. Additionally, to prevent overconfidence, we incorporate a diversity loss term. The modules in our framework complement each other, working synergistically to enhance the adaptation performance.

In summary, our paper provides the following contributions:
\begin{itemize}
  \item We propose a novel technique for estimating the reliability of PLs, called Chebyshev confidence which considers the estimated uncertainty and calculates the probability lower bound of the agreement between the model and PLs.
  \item We propose two effective denoising methods, namely direct denoising and prototypical denoising, based on the Chebyshev confidence. These methods utilize pixel and category information, respectively, for denoising.
  \item We introduce a teacher-student joint training scheme that facilitates the continuous improvement of PLs and reduces the weight of unreliable PLs.
  \item We conduct extensive experiments across various domain scenarios, including cross-centre and cross-modality settings. The results demonstrate that our model outperforms other state-of-the-art SFDA methods. The collaborative effect of our framework's modules leads to a further improvement in performance.
\end{itemize}

\section{Related Works}
\subsection{Unsupervised Domain Adaptation}
Unsupervised Domain Adaptation (UDA) has received extensive attention in the literature for visual recognition tasks \cite{guan2021domain,wang2018deep}. Previous works on UDA have utilized popular techniques such as adversarial learning \cite{wang2019boundary,hoffman2018cycada,long2018conditional,tzeng2017adversarial}, image-to-image translation \cite{hoffman2018cycada,lee2018diverse,murez2018image}, and cross-domain divergence minimization \cite{chen2020homm,abuduweili2021adaptive,shen2018wasserstein}. Self-training methods \cite{zou2018unsupervised,liu2021cycle,toldo2020unsupervised,mei2020instance} have also gained prominence in UDA, where a student model is iteratively trained using labeled source data and pseudo-labeled target data generated by a teacher model. With the increasing demand for automated medical image analysis, domain adaptation models have received considerable attention in the field of medical imaging \cite{wang2019boundary, chen2020unsupervised, xing2019adversarial}, as the procedure of manual labeling is time-consuming and requires specialized knowledge.

However, most existing UDA approaches rely on continued access to labeled source domain data during domain adaptation training, which is often impractical in real-world scenarios due to data privacy concerns. To address this limitation, the setting of SFDA has gained significant interest, as it does not require access to the source data during adaptation.

\subsection{Source-Free Domain Adaptation}
Due to concerns regarding data privacy, SFDA has emerged as an approach to achieve adaptation using only unlabeled target data and a source model, without relying on the source data. In recent years, several approaches have been proposed to address the challenges of SFDA in both natural and medical imaging domains \cite{liang2020we, wang2020tent, liu2023memory, chen2021source, yang2022source}. SHOT \cite{liang2020we} employs a centroid-based method to generate PLs for self-training and freezes the last few layers during adaptation. Tent \cite{wang2020tent} freezes parameters, except for batch normalization, and utilizes entropy minimization to update weights. OSUDA \cite{liu2023memory} applies constraints to batch normalization parameters and utilizes predicted probability for PL denoising and selection. DPL \cite{chen2021source} performs denoising on PLs at both the pixel and class levels based on uncertainty estimation. Additionally, \cite{yang2022source} leverages parameters in batch normalization to transform target images into source-style images and incorporates contrastive learning for self-training.

In SFDA, PL are widely used \cite{liang2020we, liu2023memory, chen2021source, yang2022source} but often contain noise, which can potentially mislead the model. Therefore, accurately estimating the reliability of PLs is crucial. Prior works, such as \cite{chen2021source} and \cite{litrico2023guiding}, filter out PL samples based on uncertainty. Meanwhile, \cite{liu2023memory} and \cite{xu2022denoising} primarily utilize predicted probability to assess the reliability of PLs. However, considering only probability or uncertainty alone is insufficient. In this study, we comprehensively leverage both probability and uncertainty to assess the model's confidence on PLs, enabling a more accurate denoising. Besides, we introduce a teacher-student scheme to further enhance the quality of the PLs.

\begin{figure*}
  \centering
  \includegraphics[width=\linewidth]{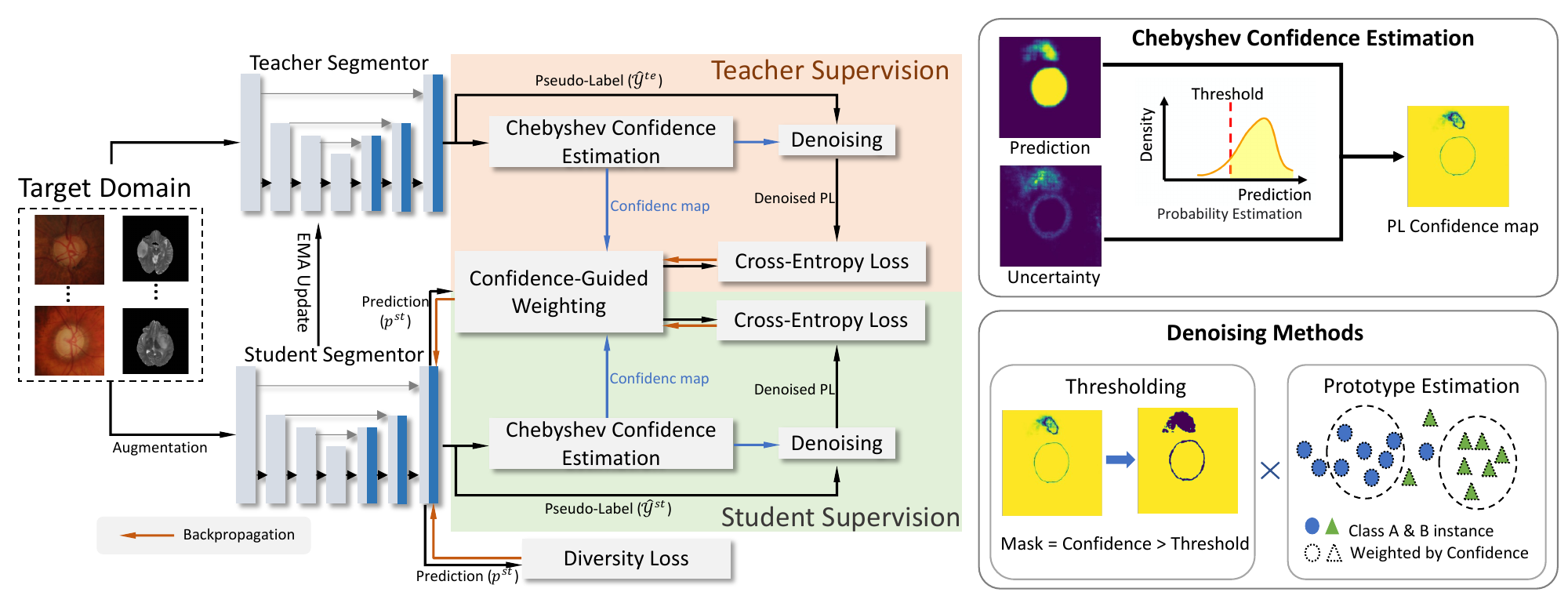}
  \captionsetup{font=scriptsize}
  \caption{Overview of our proposed framework where we use predictions from both the teacher and student models for self-training.
  We incorporate confidence guidance to assign higher weights to more accurate predictions.
  In the upper-right part of the figure, we present the Chebyshev Confidence Estimation module, which estimates the probability lower bound of the model agrees with the current pseudo-label, taking into account the model's uncertainty and predicted probability.
  Based on the estimated confidence, we apply the direct and prototypical denoising methods to refine the pseudo-labels, as shown in the lower-right part of the figure.
  }
  \label{fig:02}
\end{figure*}

\section{METHODOLOGY}
Our framework utilizes a PL-based self-training mechanism for SFDA. Given an input image, we first compute the PLs and their Chebyshev conficence scores (as introduced in Section III-B). Then, we apply a confidence-guided denoising module (as introduced in Section III-C) to remove unreliable PLs. The Teacher-Student Joint Training Scheme facilitates the iterative improvement of PLs (as introduced in Section III-D). Finally, diversity loss is incorporated to prevent overconfidence (as introduced in Section III-E).

\subsection{Problem Setting}
We denote the source domain dataset by $D_s = {\{(x_s^i, y_s^i)\}}_{i=1}^{N_s}$ where $N_s$ is the number of samples and $ y_s^i$ is the label for the image $x_s^i$. The source model, denoted as $f_{\theta_s}: x_s^i \rightarrow y_s^i$, is trained on $D_s$. The target domain dataset $D_t = {\{(x_t^i)\}}_{i=1}^{N_t}$ contains $N_t$ unlabeled samples. In unsupervised SFDA, we lack access to the target labels $ {\{(y_t^i)\}}_{i=1}^{N_t}$ throughout the entire adaptation process. Both $D_s$ and $D_t$ follow the same underlying label distribution, with a common label set $L = \{1,2,..K\}$. For the segmentation task, the medical images $x^i \in \mathbb{R}^{H\times W\times C}$ could be captured in different scenarios (e.g., fundus and brain MRI) and the corresponding label is denoted by $y^i \in \{1,0\}^{ H\times W\times K}$, where $C$ represents the number of channels in the input image, and $K$ denotes the number of classes. In this study, we focus on the SFDA problem and $D_s$ is unavailable when adapting $f_{\theta_s}$ on $D_t$.

\begin{figure}
  \centering
  \includegraphics[width=\linewidth]{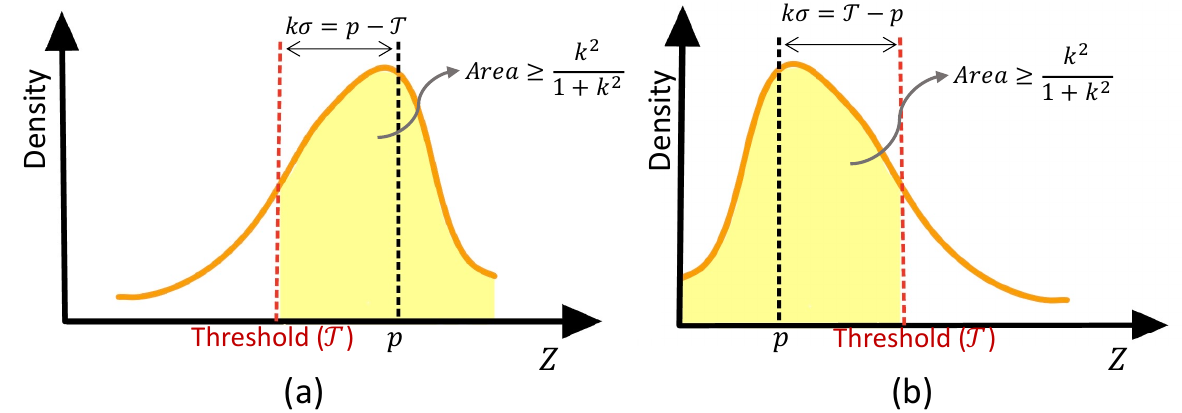}
  \captionsetup{font=scriptsize}
  \caption{Probability density function of the model's prediction for a pixel. (a) and (b) show the estimation of Chebyshev Confidence when the pseudo-label is 1 and 0, respectively.
  $p$ and $\sigma$ denote the mean predicted probability and corresponding uncertainty.
  }
  \label{fig:cheb}
\end{figure}

\subsection{Chebyshev Confidence Estimation}
In this section, we present our approach for estimating the confidence of PLs, which serves as a fundamental component of our framework.

In the PL-based method, the performance of the resulting model is directly determined by the quality of the PLs. Thus, it is crucial to assess the reliability of the PLs and guide the training procedure using only the reliable ones.
The the assessment of PL reliability cannot be circumvented in many tasks, such as denoising, estimating prototypes and evaluating the weights of PLs \cite{wang2021uncertainty,chen2021source,xu2022denoising}.
Specifically, we calculate the probability of the model predictions aligning with the current PL considering the influence of uncertainty.

Given an input image to a DNN for semantic segmentation, the corresponding output probability to a pixel $\alpha$ is denoted as $Z$. $Z$ is treated as a random variable with mean $p$ and uncertainty $\sigma$. $p$ is assigned the actual prediction of input pixel $\alpha$. To compute the uncertainty, we turn on the dropout module in the model as described in \cite{gal2016dropout} to obtain multiple samples $\{z_1,z_2,...,z_n\}$ of $Z$. Here, each element $z_i$ associates with a run of the dropout strategy.
We, then, calculate the dropout uncertainty \cite{gal2016dropout} $\sigma=\sqrt{\frac{1}{n}\sum_{n}(z_i-p)^2}$ which captures the standard deviation of $Z$.
The corresponding PL is defined as $\hat{y} = \mathbbm{1}[p\geq\mathcal{T}]$, where $\mathcal{T}\in (0,1)$ is a probability threshold to generate binary PLs. 
Ideally, we aim to estimate the probabilities $P(\mathbf{Z}>\mathcal{T}|\hat{y}=1)$ and $P(\mathbf{Z}<\mathcal{T}|\hat{y}=0)$ to evaluate the confidence of the PL.
However, directly calculating these probabilities is difficult. 
To address this, we propose a novel technique that utilizes the one-tailed variant of Chebyshev's inequality, incorporating $p$ and $\sigma$, as shown in Fig. \ref{fig:cheb}.
We first obtain the following expression for any real number $k > 0$:
\begin{equation}
  P(\mathbf{Z}-p\geq k\sigma)\leq \frac{1}{1+k^2}.
\label{eq.05}
\end{equation}
Considering $\hat{y}=0$, let $k = \frac{1}{\sigma}(\mathcal{T}-p)$.
Then, we have:
\begin{equation}
  P(\mathbf{Z}\geq \mathcal{T}|\hat{y}=0)\leq \frac{\sigma^2}{\sigma^2+(\mathcal{T}-p)^2}.
\label{eq.06}
\end{equation}
Since $P(\mathbf{Z}<\mathcal{T}|\hat{y}=0) = 1-P(\mathbf{Z}\geq \mathcal{T}|\hat{y}=0)$, we obtain
\begin{equation}
  P(\mathbf{Z} < \mathcal{T}|\hat{y}=0)\geq \frac{(\mathcal{T}-p)^2}{\sigma^2+(\mathcal{T}-p)^2}.
\label{eq.07}
\end{equation}
Similarly, for $\hat{y}=1$, we derive the following expression:
\begin{equation}
  P(\mathbf{Z} > \mathcal{T}|\hat{y}=1)\geq \frac{(p-\mathcal{T})^2}{\sigma^2+(p-\mathcal{T})^2}.
\label{eq.08}
\end{equation}
It is worth noting that the right sides of (\ref{eq.07}) and (\ref{eq.08}) estimate the lower bound of the probabilities and are actually the same. Thus, we define an indicator $l=\frac{1}{\sigma^2+(p-\mathcal{T})^2}(p-\mathcal{T})^2$ to quantify the Chebyshev confidence. The Chebyshev confidence can be easily computed given the predicted probability and uncertainty without the need for hyperparameters.

When the values of $p$ and $\mathcal{T}$ are close, the resulting PL is often unreliable, resulting in a low Chebyshev confidence.
A large $\sigma^2$ associates a small Chebyshev confindence and a high uncertainty of the model's predictions.
In summary, this Chebyshev confidence implicitly covers various scenarios and provides a concise representation, considering both the model's prediction and uncertainty. Additionally, this method can be extended to multi-class tasks by substituting $\mathcal{T}$ with the second highest predicted probability among all classes.

\subsection{Denoising Methods Based on Chebyshev Confidence}
To accurately identify noise, we propose direct denoising and prototypical denoising which leverage pixel and category information, respectively.
\subsubsection{Direct Denoising}
One common approach after obtaining the Chebyshev confidence is to apply a threshold, denoted as $\eta$, to remove points with low confidence. 
Specifically, we define a binary mask $m\in\{0,1\}$ as follows:
\begin{equation}
  m = \mathbbm{1}[l\geq\eta],
\label{eq.09}
\end{equation}
where $l$ represents the estimated Chebyshev confidence. Pixels with $l$ less than $\eta$ are considered unreliable and excluded from the loss function. Considering the initially poor quality of the PLs \cite{liu2023memory}, we linearly decrease $\eta$ from 0.99 to 0.95 during the adaptation process. 

\subsubsection{Prototypical Denoising}
Prototype estimation is a commonly used technique to capture the characteristics of the same category and guide PL refinement \cite{chen2021source, liang2020we, zhang2021prototypical}. By evaluating the consistency between the PL and the corresponding class prototype, unreliable PLs can be identified.

However, simply averaging all the PLs does not yield an accurate prototype as not all PLs are reliable.
To address this, we propose a novel confidence-weighted prototype estimation method based on Chebyshev confidence.
The prototype for class $k$ is computed as:
\begin{equation}
  z^k = \frac{\sum_{v\in\Omega}e_vl_v\mathbbm{1}[\hat{y}_v=k]}{\sum_{v\in\Omega}l_v\mathbbm{1}[\hat{y}_v=k]}.
\label{eq.10}
\end{equation}
Here, $\hat{y}_v$ is the PL for pixel $v$, and $e_v$ denotes the corresponding interpolated output feature of the backbone. 
The estimated confidence $l_v$ is used to reduce the weight of unreliable pixels. 
This approach helps mitigate the influence of outliers when computing prototypes.
$\Omega$ is a set of pixels in one mini-batch.
Based on the distance from prototypes, we compute the prototypical PL:
\begin{equation}
  \hat{y}_v^{proto} = \underset{k}{\arg\min}\|e_v - z^k\|.
\label{eq.11}
\end{equation}

Inconsistent prototypical PLs indicate a misalignment between a pixel's position in the feature space and the corresponding prediction, suggesting a higher likelihood of labeling errors. 
We remove the inconsistent prototypical PLs by updating the mask as:
\begin{equation}
 m_v = \mathbbm{1}[\hat{y}_v=\hat{y}_v^{proto}]\mathbbm{1}[l_v\geq\eta].
 \label{eq.mask}
\end{equation}
\subsection{Teacher-Stduent Joint Training Scheme}
In most existing methods, the teacher model ($f_{\theta_{te}}$) is responsible for generating PLs to guide the training of the student model ($f_{\theta_{st}}$) \cite{chen2021source,liang2020we,yang2022source}. Both $f_{\theta_{te}}$ and $f_{\theta_{st}}$ are initialized with weights $\theta_s$ trained in the source domain and $f_{\theta_{te}}$ remains fixed during the adaptation. However, due to the domain shift, the generated PLs tend to have a high error rate, which hampers the model to learn from the target domain and increases the risk of introducing erroneous knowledge.

To address this issue, we attempt to iteratively improve the accuracy of PLs during the adaptation process.
The student model adapts rapidly to the target domain through backpropagation with target domain data, leading to improved segmentation performance, as supported by previous studies \cite{lai2022decouplenet, liu2023memory, karim2023c}. In contrast, the teacher model exhibits more stable and consistent performance. Leveraging this observation, we introduce a teacher-student joint training scheme which combines PLs generated by both models for self-training. This scheme allows continuous knowledge transmission and updating between the student and teacher models, and poor knowledge can be filtered out using Chebyshev confidence.

In our framework, illustrated in Fig. \ref{fig:02}, both the teacher model ($f_{\theta_{te}}$) and the student model ($f_{\theta_{st}}$) are employed to process data from the target domain. 
The cross-entropy (CE) loss from the teacher supervision is defined as follows:
\begin{equation}
  \begin{aligned}
  \mathcal{L}_{i}^{te} = -\sum_{v}[\hat{y}_v^{te}\cdot log(p_v^{st})+(1-\hat{y}_v^{te})\cdot log(1-p_v^{st})].
\end{aligned}
\label{eq.02}
\end{equation}
Here, $\mathcal{L}_{i}^{te}$ represent the CE loss for the $i$-th sample using PLs from the teacher model.
$p_v^{st}$ represents the predicted probability of the student model for the $v$-th pixel in the $i$-th sample.
$\hat{y}_v^{te}$ is the corresponding PL generated by the teacher model.
The CE loss for the student supervision is obtained by replacing the PLs from the teacher model with the PLs from the student model, as shown below:
\begin{equation}
  \begin{aligned}
  \mathcal{L}_{i}^{st} = -\sum_{v}[\hat{y}_v^{st}\cdot log(p_v^{st})+(1-\hat{y}_v^{st})\cdot log(1-p_v^{st})],
\end{aligned}
\label{eq.02_2}
\end{equation}
where $\hat{y}_v^{st}$ denotes the PL generated by the student model.

Given the varying quality of PLs generated by the student and teacher models across different input data, we employ Chebyshev confidence weighting to assign a higher weight to PLs of superior quality. 
Inspired by \cite{long2018conditional}, 
we employ the following formula to calculate the weight of the teacher supervision:
\begin{equation}
  w_{i}^{te} = \frac{e^{-\gamma L_{i}^{te}}}{e^{-\gamma L_{i}^{te}}+e^{-\gamma L_{i}^{st}}}, \quad
  \label{eq.01}
\end{equation}
where $L_{i}^{te}=\mathbb{E}_{v\in x_i}(l_v^{te})$ and $L_{i}^{st}=\mathbb{E}_{v\in x_i}(l_v^{st})$ are the mean values of the Chebyshev confidence maps for sample $i$ from the teacher and student models, respectively.
We calculate the student weight by $w_{i}^{st} = 1-w_{i}^{te}$.
The hyperparameter $\gamma$ adjusts the effect of Chebyshev confidence.
The weighted CE loss is then built as:
\begin{equation}
  \mathcal{L}_{ce} = \sum_{i}^{N_t}w_{i}^{te}\mathcal{L}_{i}^{te} + \sum_{i}^{N_t}w_{i}^{st}\mathcal{L}_{i}^{st}.
\label{eq.03}
\end{equation}
By incorporating the denoising mask, we obtain the final CE loss:
\begin{equation}
  \mathcal{L}_{ce} = \sum_{i}^{N_t}w_{i}^{te}\sum_{v}m_v^{te}\mathcal{L}_{i,v}^{te} + \sum_{i}^{N_t}w_{i}^{st}\sum_{v}m_v^{st}\mathcal{L}_{i,v}^{st}.
\label{eq.ce_final}
\end{equation}

To continuously refine the teacher model, we utilize the exponential moving average (EMA) approach to gradually update the teacher parameters ($\theta_{te}$) using the student parameters ($\theta_{st}$). 
The update equation from iteration $j$ to $j+1$ is:
\begin{equation}
  \theta_{te}^{j+1} = \beta\theta_{te}^{j}+(1-\beta)\theta_{st}^{j},
\label{eq.04}
\end{equation}
where $\beta$ is a smoothing factor that controls the degree of change.
Additionally, our framework incorporates augmented data inputs for the student model to enhance its generalization capabilities.

\subsection{Diversity Loss}
Self-training methods may blindly trust false labels and exhibit bias towards easier classes \cite{chen2022contrastive, liang2020we}. 
To mitigate this issue and maintain prediction diversity, we introduce a regularization term in the loss function:
\begin{equation}
  \mathcal{L}_{div} = \sum_{k}\overline{p}^klog\overline{p}^k,
\label{eq.div}
\end{equation}
where $\overline{p}^k=\mathbb{E}_{x_t^i\in D_t}({p}_i^k)$ represents the average predicted probability of class $k$ for the entire target domain.
This regularization term helps to prevent overconfidence on certain predictions and promotes a more balanced and diverse output.
This diversity loss term is combined with the cross-entropy loss, resulting in the overall training loss as follows:
\begin{equation}
  \mathcal{L} = \mathcal{L}_{ce} + \lambda\mathcal{L}_{div},
\label{eq.final_loss}
\end{equation}
where $\lambda$ is a hyperparameter that balances the weight of diversity loss.

\begin{table*}[h]
\centering
\captionsetup{font=scriptsize}
\caption{Performance comparison was conducted between our method and state-of-the-art domain adaptation techniques for optic disc and cup segmentation. The evaluation metrics used were Dice coefficient (\%) and Average Surface Distance (ASD) in pixels. The best and second best results obtained are highlighted and underlined. }
\label{tab:fundus}
\begin{tabular}{lccccccccccc}
\toprule
\multirow{3}{*}{Methods} & \multirow{3}{*}{\begin{tabular}[x]{@{}c@{}}Access\\source\end{tabular} } & \multicolumn{4}{c}{RIM-ONE-r3}&\multicolumn{4}{c}{Drishti-GS}& \multirow{3}{*}{Avg. Dice$\uparrow$}& \multirow{3}{*}{Avg. ASD$\downarrow$}\\
\cmidrule(lr){3-6}\cmidrule(lr){7-10}
  & & \multicolumn{2}{c}{Optic Disc} &\multicolumn{2}{c}{Optic Cup}& \multicolumn{2}{c}{Optic Disc} & \multicolumn{2}{c}{Optic Cup}& & \\ 
\cmidrule(lr){3-4} \cmidrule(lr){5-6} \cmidrule(lr){7-8} \cmidrule(lr){9-10}
& & Dice$\uparrow$ & ASD$\downarrow$ & Dice$\uparrow$ & ASD$\downarrow$& Dice$\uparrow$ & ASD$\downarrow$ & Dice$\uparrow$ & ASD$\downarrow$& & \\
\midrule
Source only & & 83.18 & 24.15 & 74.51 & 14.44& 93.84 & 9.05 & 83.36 & 11.39 & 83.72	& 14.76 \\
Target only & $\checkmark$ & 96.8 & - & 85.6 & - & 97.4 & - & 90.1 & -& 92.48	& -\\
\midrule
BEAL \cite{wang2019boundary} & $\checkmark$ & 89.8 & - & \textbf{81.0} & - & 96.1 & - & \underline{86.2} & -  & 88.28	& -\\
Tent \cite{wang2020tent}& & \underline{90.61} & \underline{8.54} & 79.43 & \textbf{7.97}  & \underline{96.41} & \underline{4.05} & 80.77 & 12.81& 86.81	& \underline{8.34}\\
DPL \cite{chen2021source} & & 90.13 & 9.43 & 79.78 & 9.01& 96.39 & 4.08 & 83.53 & 11.39 & \underline{87.46}	& 8.48\\
FSM \cite{yang2022source} & & 87.90 & 13.06 & 79.43 & 8.32 & 95.68 & 5.52 & 82.94 & \underline{11.06} & 86.49	& 9.40\\
OSUAD \cite{liu2023memory} & & 88.61 & 10.48 & 78.44 & 8.89 & 95.99 & 4.44 & 82.33 & 11.72& 86.34	& 8.88\\
Ours & & \textbf{94.45} & \textbf{4.80} & \underline{79.85} & \underline{8.29} & \textbf{96.52} & \textbf{4.04} & \textbf{86.30} & \textbf{8.95}& \textbf{89.28}	& \textbf{6.52}\\
\bottomrule
\end{tabular}
\end{table*}

\begin{figure*}
\centering
\includegraphics[width=\linewidth]{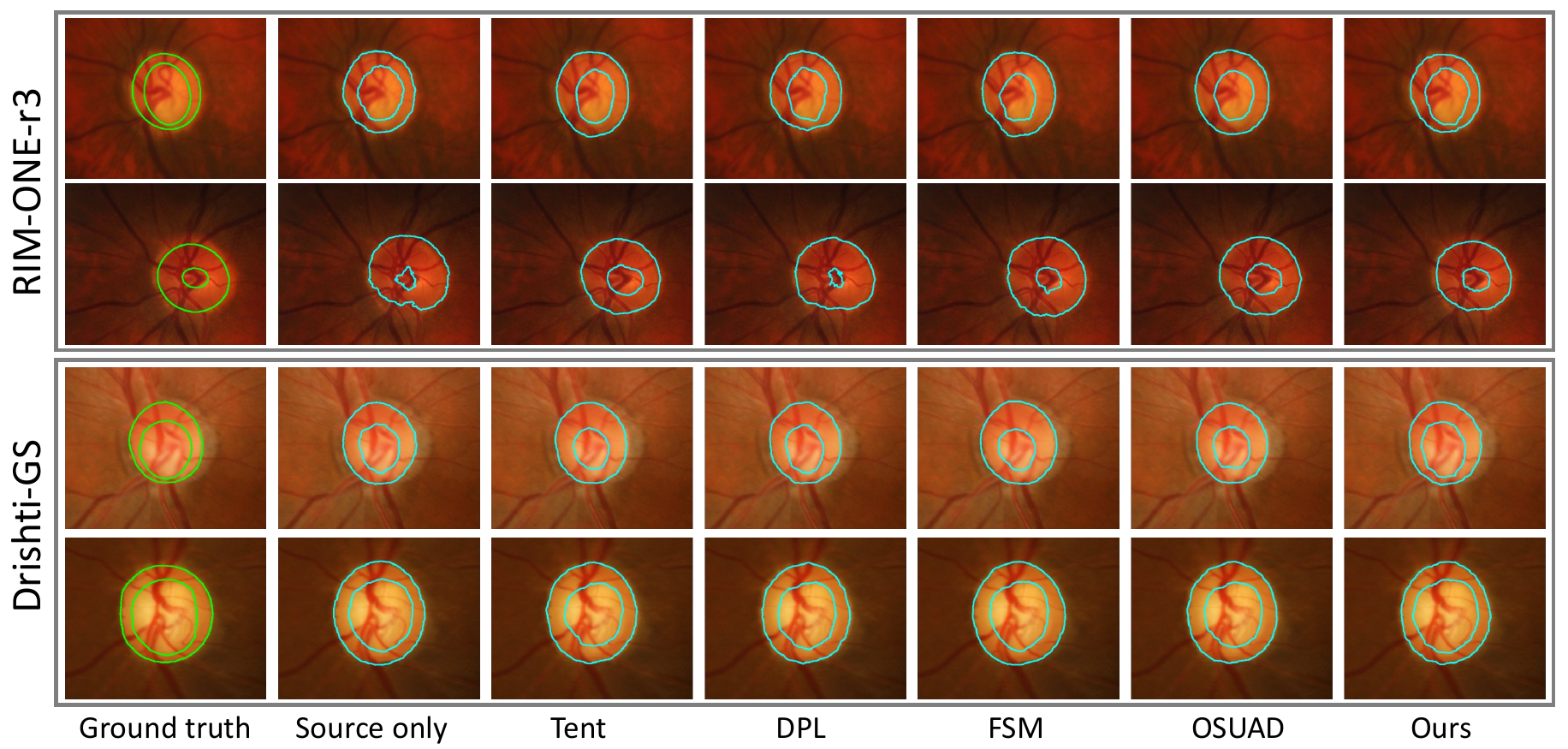}
\captionsetup{font=scriptsize}
\caption{Qualitative comparison of the adaptation performance using different methods for optic disc and cup segmentation.}
\label{fig:fundus}
\end{figure*}
  
\section{Experiments And Results}

\subsection{Datasets}
Comprehensive experiments were conducted on both fundus and brain MRI images segmentation to evaluate the proposed SFDA framework, encompassing cross-centre and cross-modality domain scenarios.

\subsubsection{Fundus Image Segmentation}
We performed SFDA for optic disc and cup segmentation of retinal fundus images using datasets from different clinical centres.
Our source domain comprised 400 annotated training images from the REFUGE challenge \cite{orlando2020refuge}. 
As distinct target domains, we utilized the RIM-ONE-r3 \cite{fumero2011rim} and Drishti-GS \cite{sivaswamy2015comprehensive} datasets.
Following the experimental protocol outlined in \cite{chen2021source}, 
the target domains consisted of training/testing subsets of 99/60 and 50/51 images, respectively. 
All images were cropped to a 512×512 disc region, which served as the input for our models.

\subsubsection{Brain Tumor Segmentation}
We conducted cross-modality SFDA on the BraTS2020 dataset \cite{bakas2018identifying}, with a specific focus on whole tumor segmentation using T1, T1ce, T2, and FLAIR modalities. 
The image volumes in BraTS originally had different resolutions, and were co-registered and interpolated to a standard resolution.
We targeted low-grade glioma cases and performed cross-modality SFDA on two pairs of MRI modalities: FLAIR$\leftrightarrow$T2 and T1$\leftrightarrow$T1ce, which exhibit relatively small appearance discrepancies. 
The target domains consists of randomly split training/testing subsets, with 53/23 cases and corresponding 3439/1487 slices. 
Each slice was resized to 512 × 512.

\subsection{Implementation Details}
Given the target images $D_t$ and the source model $\theta_s$, the student and teacher models were initialized with $\theta_s$. Subsequently, data from $D_t$ were inputted into the models, and PLs were generated. Denoising techniques were then applied to the PLs, which were used to train the student model. Our network backbone was based on a MobileNetV2 adapted DeepLabv3+ architecture \cite{chen2018encoder}, which was pretrained on ImageNet \cite{deng2009imagenet}. During the training of the source model, the segmentation network was first trained on labeled source data for 100 epochs. We used the Adam optimizer with a learning rate of 1e-3 and momentum of 0.9 and 0.99. In the adaptation stage, the model was also trained using the Adam optimizer with the same momentum. Following the approach in \cite{chen2021source}, for fundus image segmentation, the target model was trained for 2 epochs with a batch size of 8. For brain tumor segmentation, training was performed for 10 epochs to ensure convergence. The learning rate was set to be $5e^{-4}$ and $1e^{-5}$ for the fundus and brain datasets, respectively. To estimate uncertainty using Monte Carlo Dropout, we set the dropout rate to 0.5 and performed 10 stochastic forward passes to obtain the standard deviation of the output probabilities. The threshold $\mathcal{T}$ was set to 0.75, following \cite{chen2021source}. During training, weak augmentations such as Gaussian noise, Gibbs noise, contrast adjustment, and random erasing were applied to slightly perturb the input data. The hyperparameters $\gamma$, $\lambda$, and $\beta$ were set to be 1000, 0.3, and 0.999, respectively. The implementation was based on PyTorch 1.8.2 and utilized an NVIDIA V100 GPU.

For evaluation, we used two commonly-used metrics: the Dice coefficient and the Average Surface Distance (ASD). The Dice coefficient measured the overlap between the predicted segmentation and the ground truth, with higher values indicating better segmentation accuracy. The ASD measured the average distance between the predicted and ground truth surfaces, with lower values indicating better segmentation accuracy.

\subsection{Competitive State-of-the-Art Domain Adaptation Methods}
To assess the effectiveness of our proposed SFDA method, we conducted a comparison with several state-of-the-art methods, namely BEAL \cite{wang2019boundary}, Tent \cite{wang2020tent}, DPL \cite{chen2021source}, FSM \cite{yang2022source}, and OSUAD \cite{liu2023memory}. 
The method proposed by BEAL \cite{wang2019boundary} is an unsupervised domain adaptation (UDA) approach designed specifically for the fundus dataset, utilizing both source and target images during adaptation. 
On the other hand, DPL \cite{chen2021source}, FSM \cite{yang2022source} and OSUAD \cite{liu2023memory} are the latest SFDA methods that focus on the medical images. These SFDA methods are provided with a source model but not the source data for target domain adaptation. We also compare the SFDA mode of Tent \cite{wang2020tent}, which primarily focuses on natural datasets. Given that these methods are trained on different tasks with diverse datasets, it is not appropriate to directly apply them to our evaluation. Therefore, to ensure a fair comparison, we trained SFDA methods using the same MobileNetV2 backbone \cite{sandler2018mobilenetv2}. Additionally, we included the Source only and Target only models, which were trained exclusively on either the source or target data. These models serve as the lower and upper bounds for the domain adaptation problem, respectively.

\begin{table*}[htbp]
\centering
\captionsetup{font=scriptsize}
\caption{Performance Comparison with State-of-the-Art Domain Adaptation Methods on Brain Tumor Segmentation using Dice (\%) and ASD (pixel). Best and second-best results are highlighted and underlined.}
\label{tab:brats}
\begin{tabular}{lccccccccccc}
\toprule
\multirow{2}{*}{Methods}& \multirow{2}{*}{\begin{tabular}[x]{@{}c@{}}Access\\source\end{tabular} }& \multicolumn{2}{c}{T2 $\to$ FlAIR} & \multicolumn{2}{c}{FLAIR $\to$ T2} & \multicolumn{2}{c}{T1 $\to$ T1CE} & \multicolumn{2}{c}{T1CE $\to$ T1} & \multirow{2}{*}{Avg. Dice$\uparrow$} & \multirow{2}{*}{Avg. ASD$\downarrow$}\\
\cmidrule(lr){3-4}\cmidrule(lr){5-6}\cmidrule(lr){7-8}\cmidrule(lr){9-10}
& & Dice$\uparrow$ & ASD$\downarrow$ & Dice$\uparrow$ & ASD$\downarrow$ & Dice$\uparrow$ & ASD$\downarrow$ & Dice$\uparrow$ & ASD$\downarrow$ & &  \\
\midrule
Source only     &             & 62.84 & 20.15 & 73.38 & 12.77 & 66.30 & 15.98 & 71.98 & 14.79 & 68.63 & 15.92 \\
Target only     & \checkmark  & 88.30 & 4.53 & 88.78 & 5.08 & 76.48 & 9.38 & 78.47 & 8.76 & 83.01 & 6.93 \\
\midrule
Tent \cite{wang2020tent}  &             & 69.55 & \underline{11.91} & 74.06 & 12.72 & 72.99 & 11.41 & 74.93 & 10.10 & 72.88 & 11.54 \\
DPL \cite{chen2021source} &             & 69.35 & 12.13 & 74.40 & 11.97 & 72.09 & 11.28 & 73.81 & 10.67 & 72.41 & 11.51 \\
FSM \cite{yang2022source}   &             & 70.00 & 13.06 & 74.70 & 12.27 & 72.10 & 11.55 & 73.87 & 10.70 & 72.67 & 11.89 \\
OSUAD \cite{liu2023memory}  &             & \underline{73.69} & 12.26 & \underline{77.06} & \underline{11.91} & \underline{75.26} & \underline{10.32} & \underline{77.19} & \textbf{9.43} & \underline{75.80} & \underline{10.98} \\
Ours            &             & \textbf{74.82}  & \textbf{11.22} & \textbf{78.89} & \textbf{11.71} & \textbf{76.43} & \textbf{10.02} & \textbf{78.28}& \underline{9.44} & \textbf{77.10} & \textbf{10.60} \\
\bottomrule
\end{tabular}
\end{table*}

\begin{figure*}
\centering
\includegraphics[width=\linewidth]{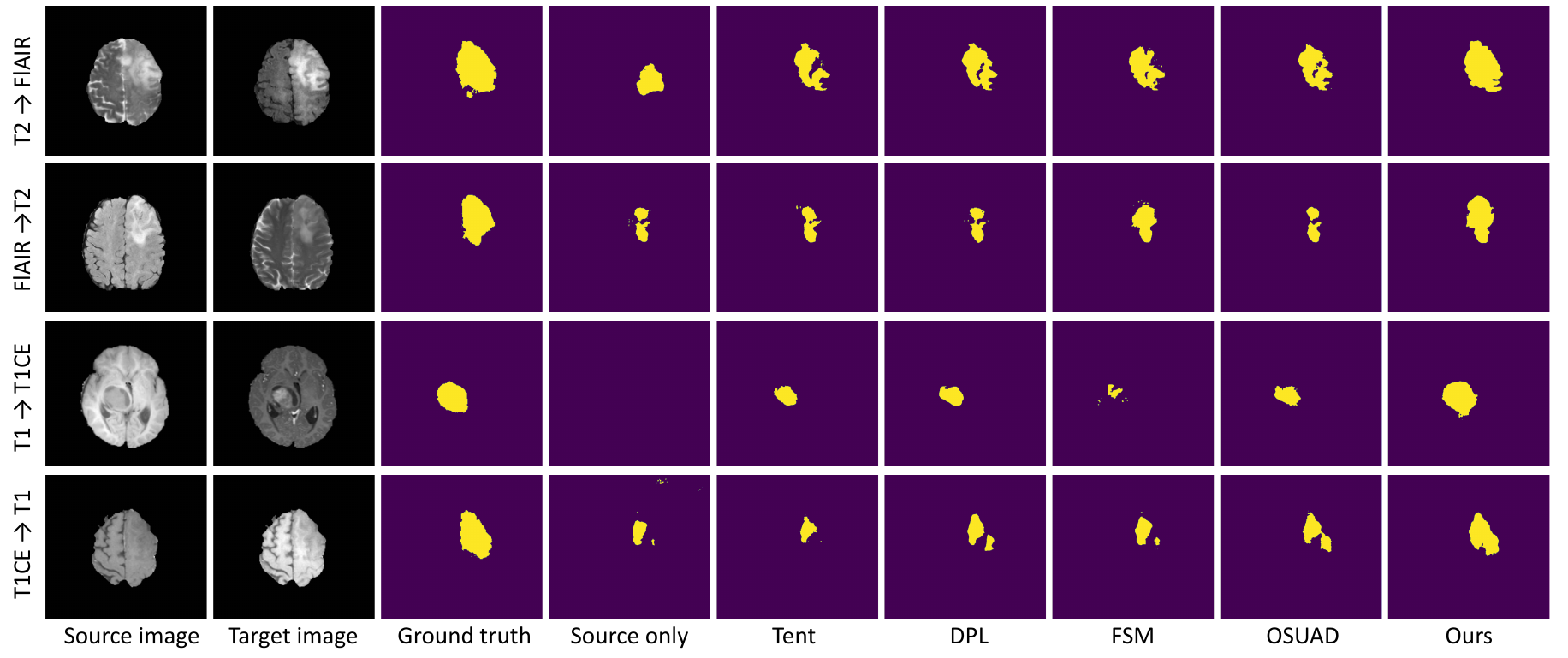}
\captionsetup{font=scriptsize}
\caption{Qualitative comparison of adaptation performance with different methods on brain tumor segmentation.
The input of the adapted model is target domain images, and the first column shows the corresponding source domain images.}
\label{fig:Brats}
\end{figure*}

\subsection{Experimental Results}

\subsubsection{Results on Optic Disc and Cup Segmentation}
Table \ref{tab:fundus} presents a performance comparison of Optic Disc and Cup Segmentation using Dice and ASD metrics on two target datasets: RIM-ONE-r3 and Drishti-GS. We refer to the results reported in the BEAL paper \cite{wang2019boundary}, which developed an unsupervised domain adaptation (UDA) model for cross-domain fundus image segmentation. Additionally, we incorporate the findings from DPL \cite{chen2021source}, which employed the same experimental setting as ours. We retrained other SOTA methods using the same backbone.
Comparing the performance of the models, our approach achieves a significant improvement in Disc segmentation, surpassing other methods by 3.84\% and 3.74 in Dice and ASD metrics, respectively, for the RIM-ONE-r3 dataset. 
On Drishti-GS, our method demonstrates the best performance for both Disc and Cup segmentation.
Specifically, the Cup segmentation results show a notable enhancement with a 2.77\% increase in Dice compared to other SFDA methods.
The improvement in Disc segmentation is minimal due to its proximity to the upper bound, represented by the target-only model.

Overall, our approach outperforms all other methods on average, including the UDA model, highlighting its effectiveness in achieving high segmentation accuracy for Optic Disc and Cup.
These improvements can primarily be attributed to the utilization of a teacher-student scheme, which facilitates the generation of more accurate PLs.
Additionally, the application of the proposed denoising method is crucial in ensuring model stability and preventing the acquisition of detrimental knowledge.

To further validate the qualitative effectiveness of our proposed method, we provide visualizations of the segmentation in Figure \ref{fig:fundus}. The first, third, and fourth rows demonstrate significant improvement, particularly in Cup segmentation. The second row showcases improvements in both Disc and Cup segmentation.

\begin{table}[htbp]
\centering
\captionsetup{font=scriptsize}
\caption{Ablation study with different experimental settings on the fundus and BraTS datasets. This table presents the average performance of all scenarios on both datasets.}
\label{tab:ablation}
\resizebox{\linewidth}{!}{%
\begin{tabular}{ccccccccc}
\toprule
\multirow{2}{*}{\begin{tabular}[x]{@{}c@{}}Diversity\\Loss\end{tabular}}& \multirow{2}{*}{\begin{tabular}[x]{@{}c@{}}Student\\Branch\end{tabular}} & \multirow{2}{*}{\begin{tabular}[x]{@{}c@{}}Confidence\\Weighting\end{tabular}} & \multirow{2}{*}{\begin{tabular}[x]{@{}c@{}}Direct\\Denoising\end{tabular}} & \multirow{2}{*}{\begin{tabular}[x]{@{}c@{}}Prototypical\\Denoising\end{tabular}} & \multicolumn{2}{c}{Fundus Dataset} & \multicolumn{2}{c}{BraTS Dataset} \\
\cmidrule(lr){6-7}\cmidrule(lr){8-9}
  &   &  &  &  & Dice$\uparrow$ & ASD$\downarrow$ & Dice$\uparrow$ & ASD$\downarrow$ \\
\midrule
  &   &  &  &  & 86.43 & 8.84 & 70.20 & 12.36 \\
\checkmark&   &  &  &  & 86.48 & 9.04 & 72.41 & 12.24 \\
\checkmark& \checkmark   &  &  &  & 86.93 & 8.61 & 71.20 & 11.57 \\
\checkmark & \checkmark & \checkmark &  &  & 87.78 & 7.65 & 71.51 & 11.63 \\
\checkmark & \checkmark & \checkmark & \checkmark &  & 88.55 & 7.12 & 75.56 & 11.21 \\
\checkmark & \checkmark & \checkmark &  & \checkmark & 88.64 & \underline{6.75} & 75.84 & 10.74 \\
  & \checkmark & \checkmark & \checkmark  & \checkmark & \underline{88.69} & 6.91 & 73.76 & 11.07 \\
\checkmark  &  &  & \checkmark & \checkmark & 87.66 & 7.37 & 76.62 & \underline{10.65} \\
\checkmark & \checkmark &  & \checkmark & \checkmark & 87.93 & 7.11 & \underline{76.90} & 10.77 \\
\checkmark & \checkmark & \checkmark & \checkmark & \checkmark & \textbf{89.28} & \textbf{6.52} & \textbf{77.10} & \textbf{10.60} \\
\bottomrule
\end{tabular}%
}
\end{table}

\begin{figure}
\centering
\captionsetup{font=scriptsize}
\includegraphics[width=\linewidth]{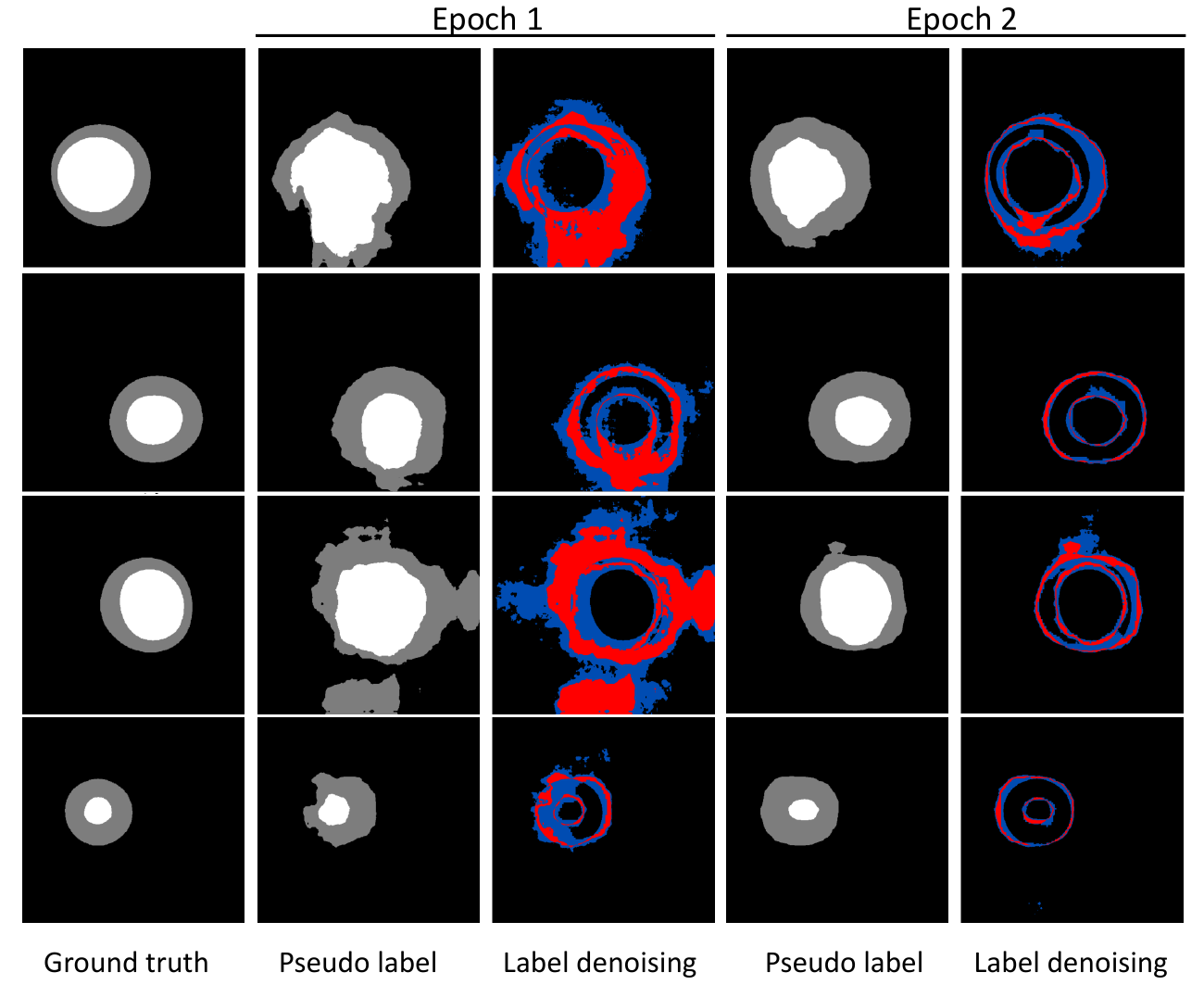}
\caption{Examples of pseudo-labels with higher weight during adaptation for fundus images. 
The correctly and falsely identified noisy pseudo-labels are indicated in red and blue colors. }
\label{fig:denoise}
\end{figure}

\subsubsection{Results on Brain Tumor Segmentation}
Table \ref{tab:brats} presents a comprehensive performance comparison for brain tumor segmentation. 
We retrained all SOTA methods using the experimental setting.
A comprehensive performance comparison for brain tumor segmentation is presented in Table \ref{tab:brats}.
Our model consistently achieves the highest Dice values across all scenarios. In most cases, it also achieves the lowest ASD values, except for T1CE $\to$ T1, where it performs as the second-best.
On average, our model outperforms all other models, demonstrating an improvement of 8.47\% in Dice and 5.32 in ASD compared to the source-only model. 
These results emphasize the strong performance of our method compared to other approaches.
The notable improvements in our model can be attributed to the utilization of an enhanced denoising method and iteratively improved PL. 
Moreover, the incorporation of a diversity loss term plays a crucial role in preventing overconfidence in the model's predictions.

To qualitatively validate the effectiveness of our method, we visualize the segmentation predictions in different scenarios as shown in Figure \ref{fig:Brats}. In these samples, other methods generally demonstrate poor performance with numerous false negatives, whereas our method consistently produces more accurate brain tumor segmentations. Besides, our segmented regions closely resemble regular tumors with no hollow areas or outliers.
In addition, the corresponding source images are displayed in the first column, revealing the evident change in object intensity between the source and target images, which highlights the capability of our method.

\begin{figure}
\centering
\includegraphics[width=0.85\linewidth]{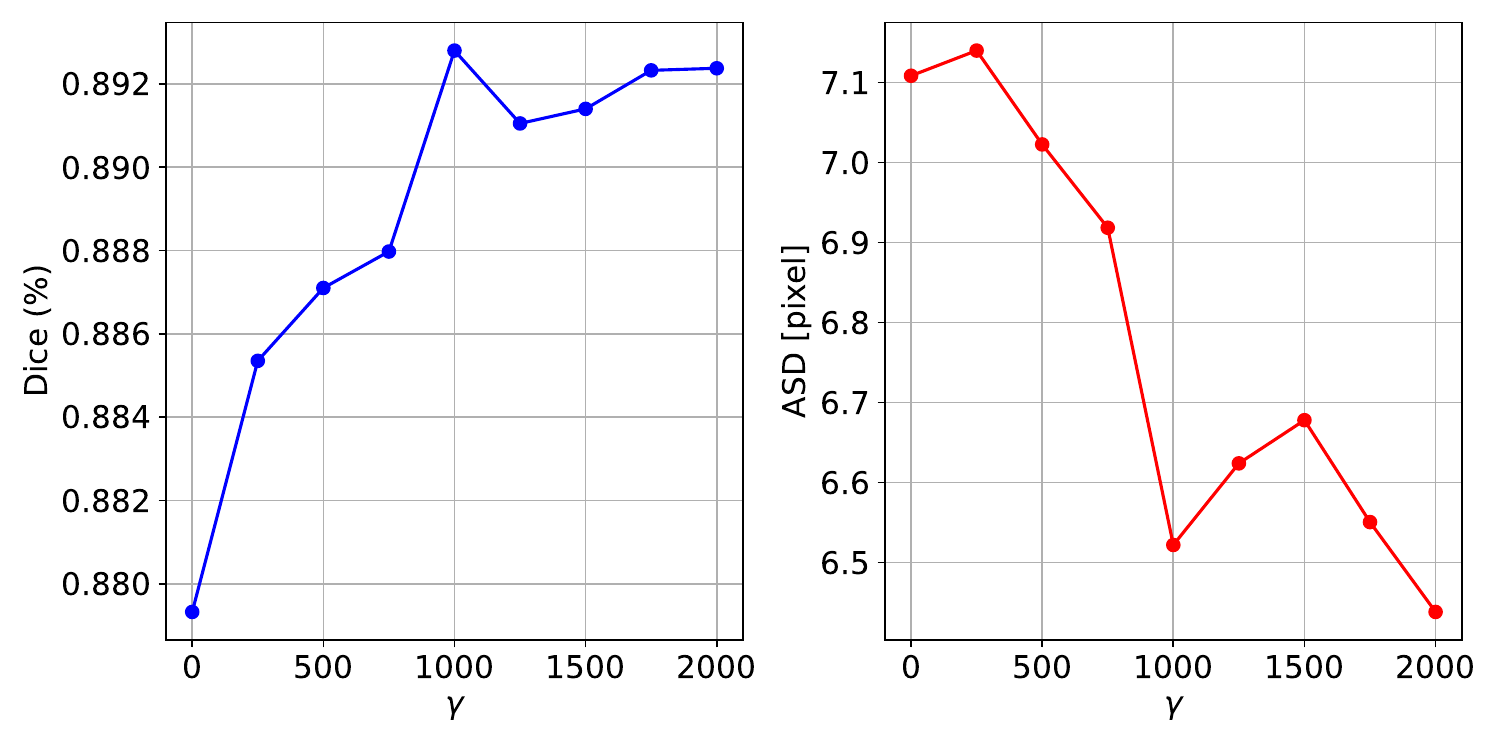}
\captionsetup{font=scriptsize}
\caption{Average segmentation performance of optic disc and cup on both RIM-ONE-r3 and Drishti-GS target domains with varying $\gamma$ values of the confidence-guided weighting module.}
\label{fig:gamma}
\end{figure}

\subsection{Ablation Analysis of Key Components}
\subsubsection{Effectiveness of Teacher-Student Joint Training Scheme}
The teacher-student Scheme can be splited into a student branch module and a confidence-guided weighting module. When the confidence-guided weighting is removed, $\gamma$ is set to 0, ensuring equal weighting between the student and teacher branches.

For the fundus dataset, Table \ref{tab:ablation} (rows 3, 4, 9, and 10) demonstrates the performance enhancement achieved by incorporating the student branch into the model, and the addition of confidence-guided weighting further improves the model's performance. While the direct inclusion of the student branch yields performance improvements, confidence-guided weighting remains critical. It enables the model to select high-quality PLs for learning and suppresses the impact caused by poor-quality PLs. Figure \ref{fig:gamma} illustrates the variations in segmentation performance on the fundus dataset for different $\gamma$ values. The graph shows that the segmentation performance continuously improves before $\gamma$ reaches 1000. Subsequently, further increases in $\gamma$ do not significantly boost the model's performance.
This is because when $\gamma$ is small, the weight difference is small, thus the effect is not significant. This is because when $\gamma$ is small, the weight difference is small, thus the effect of the scheme is not significant.

For the BraTs dataset, the teacher-student scheme enhances the model's performance (0.58\% improvement in Dice) when combined with denoising.
However, incorporating the teacher-student scheme without denoising yields negligible improvements. 
This can be attributed to the higher level of noise present in the modality adaptation within the BraTS dataset. 
Additionally, the results in rows 9 and 10 of Table \ref{tab:ablation} indicate that the BraTS dataset results are not sensitive to changes in $\gamma$.

Fig. \ref{fig:denoise} displays the PLs generated by the student branch at different adaptation stages.
The quality of the PLs progressively improves throughout the adaptation process, accompanied by a reduction in the denoised area. 
This iterative refinement provides the model with more accurate and fine domain knowledge.

\begin{table}[htbp]
\centering
\captionsetup{font=scriptsize}
\caption{Average performance of different denoising methods on all scenarios. To calculate the following results, we set the threshold for entropy, uncertainty, and Chebyshev confidence to 0.1, 0.05, and 0.05, respectively.}
\label{tab:denoising}
\begin{tabular}{cccc}
\hline
Method & F1 score$\uparrow$& Accuracy$\uparrow$ \\
\hline
Entropy & \underline{0.3861} $\pm$ 0.0633 & $0.9568 \pm 0.0194$ \\
Uncertainty & $0.2704 \pm 0.1861$ & $0.9741 \pm 0.0180$ \\
Prototypical denoising & $0.2739 \pm 0.1665$ & \textbf{0.9796} $\pm$ 0.0052 \\
DPL mask & $0.3721 \pm 0.0836$ & $0.9731 \pm 0.0136$ \\
Our mask & \textbf{0.4054} $\pm$ 0.0606 & \underline{0.9767} $\pm$ 0.0083 \\
\hline
\end{tabular}
\end{table}

\begin{figure}
\centering
\includegraphics[width=\linewidth]{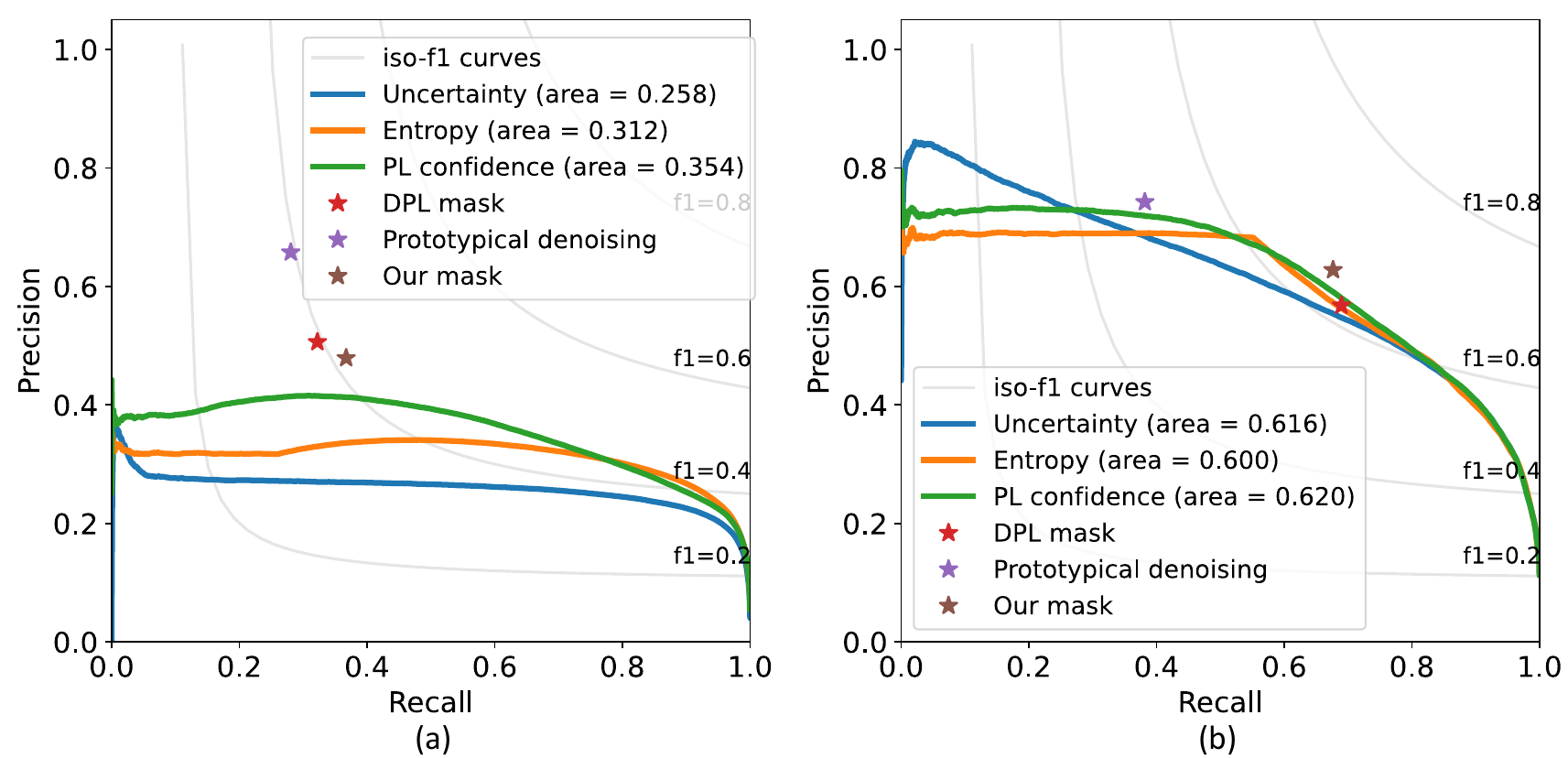}
\captionsetup{font=scriptsize}
\caption{(a) and (b) depict the precision and recall curves for the denoising performance of fundus and brain images, respectively.}
\label{fig:PR}
\end{figure}

\subsubsection{Effectiveness of Proposed Denoising Method}
Comparing the 2rd and 8th rows, as well as the 4th and 10th rows, of Table \ref{tab:ablation} reveals a significant performance improvement by incorporating the proposed denoising method. 
Specifically, there is an increase of approximately 1.83\% and 4.90\% in the Dice coefficient on average for the fundus and BraTs datasets, respectively.
Notably, the denoising method shows greater significance on the BraTs dataset, possibly due to the larger domain shift that necessitates noise reduction in PLs.

The 5th and 6th rows demonstrate that incorporating either direct denoising or prototypical denoising individually leads to performance enhancements. 
The simultaneous utilization of both denoising techniques further amplifies the model's performance for both the fundus and BraTs datasets. 
This finding suggests a certain degree of complementarity between direct denoising and prototypical denoising, and their simultaneous use yields better denoising outcomes.

\subsubsection{Comparison of Denoising Methods}
To further evaluate our proposed denoising method, we compared it with commonly used denoising methods, including entropy, uncertainty, and the DPL mask \cite{chen2021source}. 
Denoising can be seen as a binary classification task, where the objective is to classify correct PLs from incorrect ones. 
Hence, evaluation metrics from binary classification can be employed.

Table \ref{tab:denoising} showcases the performance of different denoising methods based on F1 score and Accuracy. It can be observed that Our mask achieves the highest F1 score, surpassing the second-best method by 1.93\%. Furthermore, Our mask achieves a comparable accuracy to the best method, with only a marginal difference of 0.29\%. The high F1 score and accuracy indicate that our denoising method performs well in classifying both noisy and non-noisy regions.

The precision and recall curves shown in Fig. \ref{fig:PR} were generated by randomly selecting 16 images from each dataset to evaluate different denoising methods.
From Fig. \ref{fig:PR}, it can be observed that our proposed Chebyshev confidence outperforms uncertainty and entropy in terms of classification performance, with respective area values of 0.354 and 0.620 for the fundus and BraTs datasets.
This can be attributed to the improved assessment of PL reliability provided by our designed Chebyshev confidence.
Furthermore, we can observe that prototypical denoising achieves the highest precision at the same recall value.
This highlights the superiority of the prototypical approach. 
However, it should be noted that prototypical denoising generates a binary mask that cannot be adjusted by manipulating the threshold.
Therefore, the combination of prototypical denoising and direct denoising using Chebyshev confidence provides a more flexible denoising approach. 
We believe that this is one of the reasons why prototypical denoising and direct denoising can complement each other.

\subsubsection{Effectiveness of Diversity Loss}
We also investigated the influence of diversity loss on the model's performance. The results, as shown in the second and seventh rows of Table \ref{tab:ablation}, indicate that incorporating diversity loss yields performance enhancements for both datasets. Specifically, for the fundus dataset, the model demonstrates improvements of 0.32 in Dice coefficient and 0.10 in ASD on average. In the case of the BraTs dataset, the observed improvements are more pronounced, with gains of 2.78 in Dice coefficient and 0.3 in ASD. The significant impact of diversity loss on the BraTs dataset can be attributed to its longer adaptation over 10 epochs, which may exacerbate the issue of overconfidence. In such scenarios, the inclusion of diversity loss becomes crucial as a means to prevent overconfidence and learning incorrect knowledge.

\section{Conclusion}
In this paper, we present a SFDA framework that aims to effectively address the noise present in pseudo-labels and iteratively refine them. The SFDA framework addresses privacy concerns by solely transferring knowledge from a well-trained source model to the target domain. We introduce a novel technique called Chebyshev confidence to accurately estimate the reliability of pseudo-labels, which is independent of hyper-parameters. Building upon the Chebyshev confidence, we propose two confidence-guided methods: direct denoising and prototypical denoising. These methods leverage pixel and category information, respectively, to eliminate noise from the pseudo-labels. Furthermore, we introduce a teacher-student joint training scheme that iteratively enhances the accuracy of the pseudo-labels and assigns higher priority to high-quality ones. To prevent overconfidence, we incorporate a diversity loss term. By integrating these modules, our framework generates more accurate and cleaner pseudo-labels for self-training, leading to stable adaptation and improved performance. We have conducted experiments on various scenarios, including cross-centre and cross-modality adaptation, and the results demonstrate the superiority of our framework. Our proposed framework has the potential to be widely applicable in scenarios that necessitate high-quality pseudo-labels.


\end{document}